\documentclass[letterpaper,twocolumn,10pt]{article}
\usepackage{usenix-2020-09}

\usepackage{tikz}
\usepackage{amsmath}

\usepackage{filecontents}

\usepackage{placeins} 
\usepackage{float}
\usepackage{booktabs} %
\usepackage{algorithm}
\usepackage{algpseudocode}
\usepackage{pifont}  
\usepackage{comment} 
\usepackage{enumitem}
\usepackage{subcaption}

\begin{document}

\date{}

\title{STS: Efficient Sparse Attention with \underline{S}peculative \underline{T}oken \underline{S}parsity}
\newcommand{\name}{STS }
\newcommand{\namens}{STS}
\newcommand{\namess}{STS's }
\newcommand{\citeme}{\textcolor{red}{[?]}}

\newcommand{\fixme}[1]{\textcolor{red}{#1}}

\author{
{\rm Jiangnan Yu\thanks{Both authors contributed equally.}, Ceyu Xu\footnotemark[1], Yongji Wu\thanks{Corresponding author.}, and Yuan Xie}\\
Jiangnan Yu, Ceyu Xu, and Yuan Xie: The Hong Kong University of Science and Technology\\
Yongji Wu: UC Berkeley
}

\maketitle


\begin{abstract}


The quadratic complexity of attention imposes severe memory and computational bottlenecks on Large Language Model (LLM) inference.
This challenge is particularly acute for emerging agentic applications that require processing multi-million token sequences.
We propose \namens, a sparse attention mechanism that requires no model retraining.
\name leverages the key insight that \textbf{tokens identified as important by a smaller draft model are highly predictive of important tokens for a larger target model.}
By integrating into speculative decoding frameworks, \name repurposes the draft model's attention scores to dynamically construct a token-and-head-wise sparsity mask.
This mask effectively prunes the expensive attention computation in the target LLM.
Our evaluation shows that \name{} achieves a 2.67$\times$ speedup operating at approximately 90\% sparsity on representative benchmark NarrativeQA, maintaining negligible accuracy degradation compared to dense attention.
\name establishes a new state-of-the-art on the sparsity-accuracy trade-off, outperforming prior techniques by enabling higher sparsity levels for a given accuracy budget.

\end{abstract}

\section{Introduction}
The power of the attention mechanism~\cite{vaswani2017attention} originates from its ability to construct a globally aware context by computing relevance scores between every pair of tokens in the sequence. 
This ``attend-to-everything'' approach is a double-edged sword, enabling unprecedented model capability while incurring quadratic complexity as the sequence scales: 
On the one hand, this architecture has become ubiquitous in modern AI, serving as the foundation for the emergence of advanced machine intelligence.
Conversely, the necessity of attending to all preceding tokens imposes severe computational and memory-access overheads, which become prohibitive in applications that require long sequence lengths.

\begin{figure}
    \centering
    \includegraphics[width=\columnwidth]{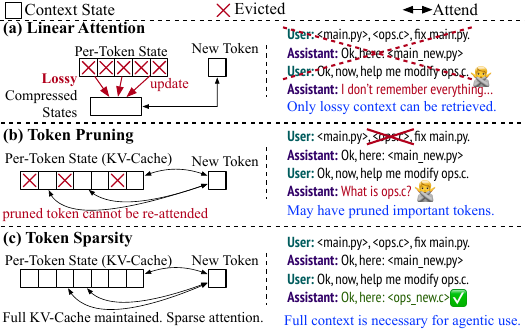}
    \caption{A taxonomy of methods for reducing the computation cost of dense attention: (a) Linear Attention, (b) Token Pruning, and (c) Token Sparsity. }
    \label{fig:linear_prune_sparsity}
\end{figure}

To mitigate these costs, prior work~\cite{yang2024tidaldecode,tang2024quest,xiao2023streamingllm,seerattention,zhang2023h2o,gu2024mamba,choromanski2020performer,kitaev2020reformer,yuan2025nsa} focuses on designing an approximation of the attention mechanism that trades a bit of accuracy for lower computational cost.
One class of approximation is linear attention~\cite{kitaev2020reformer,wang2020linformer,choromanski2020performer,gu2024mamba} or token pruning~\cite{zhang2023h2o,xiao2023streamingllm} as shown in Figure \ref{fig:linear_prune_sparsity} (a) and (b).
However, these methods often cause irreversible information loss, which is risky because the importance of a token is dynamic, so a token that is irrelevant at one step may become critical later. 
Consequently, lossy methods can cause model accuracy degradation when the model requires historical information that was previously discarded. 
A more robust alternative is \textit{Token Sparsity}~\cite{yang2024tidaldecode,tang2024quest}. 
This approach maintains the full KV cache but dynamically selects a small subset of critical tokens for the current query. 
The central question thus becomes how to design an algorithm that accurately and efficiently identifies this critical subset of tokens for the current query, without prematurely discarding the full context needed for future queries.

Accurately identifying the importance of a specific token without performing the full attention computation remains a non-trivial challenge.
Existing approaches typically rely on heuristics~\cite{yang2024tidaldecode,tang2024quest} and suffer from three fundamental limitations that impede practical deployment.
First, lightweight selection algorithms often achieve suboptimal accuracy; when critical tokens are missed, they must retain significantly more tokens than necessary to preserve model quality, or else accuracy degrades.
Second, many methods reduce the search space by partitioning tokens into blocks and selecting entire blocks rather than individual tokens, which limits the maximum achievable sparsity.
Third, the token selection process incurs computational overhead that increases LLM inference latency, often offsetting the performance gains from reduced attention complexity.

In this paper, we address these challenges with \underline{S}peculative \underline{T}oken \underline{S}parsity (\namens), a sparse attention algorithm that is accurate, efficient, and supports per-token granularity by leveraging the speculative decoding framework.
The efficacy of \name stems from a key observation: for a given input, the attention patterns of two different models exhibit strong correlation, especially when the models belong to the same family. 
Leveraging this insight, \name repurposes the small draft model so that, in addition to its standard role of proposing new candidate tokens, it now also generates a dynamic sparsity mask from its own attention computations. The large target model then uses this mask to sparsify its attention computation. 
Consequently, the prohibitively expensive dense attention is confined to the small draft model. 
In contrast, the target model performs a highly efficient sparse attention operation, drastically reducing both computational and memory bandwidth requirements with minimal impact on model accuracy.

Deriving the sparsity mask from the draft model offers several advantages. 
First, \name is training-free, simplifying its deployment. 
Second, its seamless integration with speculative decoding frameworks makes it well-suited for low-latency inference. \name benefits from reduced attention computation due to sparsification, and unlike prior approaches that perform token selection immediately before each attention computation, \name executes token selection on the smaller draft model asynchronously with respect to the target model's execution path. Because this selection finishes before the target model needs the sparsity masks, its latency can be hidden within the main execution pipeline.
In addition, this ``known-in-advance'' property distinguishes \name from just-in-time mask generation methods, which enables further optimizations, such as proactive data prefetching to hide communication latency in a KV-cache offload scenario.
Moreover, while existing methods often revert to dense computation during the prefill phase, negating performance gains in multi-turn agentic interactions~\cite{liu2024agentbench,wang2024mint}, \name maintains sparsity across both prefill and decode phases.
Finally, compared to existing token sparsity schemes, \name achieves a state-of-the-art accuracy-sparsity trade-off, demonstrating that the draft model serves as a highly effective proxy for token selection.

This paper makes the following contributions:
\begin{enumerate}
  \item We propose \namens, a training-free sparse attention mechanism applicable to both prefill and decode stages. It significantly reduces the memory bandwidth and computational requirements of LLMs with negligible impact on model accuracy.
  \item We implement \name in vLLM~\cite{kwon2023efficient}, demonstrating its seamless integration into speculative decoding frameworks. Our evaluation shows that \name{} achieves a 2.67$\times$ speedup operating at approximately 90\% sparsity on representative benchmark NarrativeQA, maintaining negligible accuracy degradation compared to dense attention.
  \item We evaluate \name against state-of-the-art token sparsity techniques and show that it achieves a superior accuracy-sparsity trade-off, enabling higher sparsity for a given accuracy budget (and vice versa).
  \item We identify and analyze the ``known-in-advance'' property of \namens, where the sparsity mask is determined before the target model's computation. We show that this unique feature enables further latency optimizations and opens new avenues for future, more clever memory orchestrations.
\end{enumerate}

\section{Background}
This section first provides context on speculative decoding. We then discuss the taxonomy of attention approximation techniques, comparing their respective advantages and disadvantages to contextualize our proposed approach.

\subsection{Speculative Decoding}
The autoregressive nature of large language models (LLMs) makes generation a sequential process, in which each model generates only one new token at a time. 
For each new token, the model must scan its entire set of parameters and the complete Key-Value (KV) cache of all preceding tokens. 
This makes the process extremely memory-intensive, causing overall decoding latency to be largely bottlenecked by GPU memory bandwidth.

Speculative  Decoding~\cite{leviathan2023fast,chen2023accelerating} aims to solve this issue by breaking the inherent sequentiality, enabling the LLM to decode multiple tokens in a single forward pass. 
It achieves this by using a smaller, faster "draft" model to generate a sequence of new candidate tokens, while the larger "target" model then behaves like a verifier, checking all candidates in a single parallel decoding step. 
The reason this is effective is that although the draft models are smaller and less capable, they can still generate a relatively correct sequence, especially when some tokens are not that hard to predict. 
In a successful speculation, a prefix of candidate tokens is accepted, allowing the model to take multiple steps in a single forward pass and achieving a significant speedup. 
When a prediction fails, forward progress is still guaranteed. 
The system accepts correctly guessed tokens up to the point of the mismatch, then appends the single correct token produced by the target model for that position.
In reality, the overhead of the draft model is so small that it becomes almost free, resulting in pure latency reduction.

The core idea of speculative decoding, therefore, is to use a small model to guide a large model. In our work, STS, we learn from this principle and extend it, positing that the small model's guidance can be applied not only to predicting output tokens but also to approximating the expensive internal computations of the large model.


\begin{table}[ht]
\caption{A comparison of methods to reduce the computational cost of dense attention. (St. LLM stands for StreamingLLM~\cite{xiao2023streamingllm} and TidalDec. stands for TidalDecode~\cite{yang2024tidaldecode} and Lin. Attn. stands for Linear Attention~\cite{wang2020linformer})}
\centering
\footnotesize
\begin{tabular}{|c|c|c|c|c|c|c|c|}
\hline
\textbf{Method} & \textbf{Lin. } & \textbf{St.} & \textbf{H2O} & \textbf{Quest} & \textbf{Tid.} & \textbf{NSA} & \textbf{STS*} \\ 
 & \textbf{Att.} & \textbf{LLM} & \cite{zhang2023h2o} & \cite{tang2024quest} & \textbf{Dec.} & \cite{yuan2025nsa} &  \\
\hline
\hline
Lossless & N & N & N & Y & Y & Y & Y \\
Context &  &  &  &  &  &  &  \\
\hline
Sparse & - & Y & N & N & N & Y & Y \\
Prefill &   &   &   &   &   &   &   \\
\hline
Sparse & - & Y & Y & Y & Y/N & Y & Y \\
Decode &   &   &   &   &  &  &  \\
\hline
No Extra & - & Y & Y & N & N & N & Y \\
Latency &  &  &  &  &  &  &  \\
\hline
\end{tabular}
\label{table:method_comparison}
\end{table}







\begin{figure*}
    \centering
    \includegraphics[width=\textwidth]{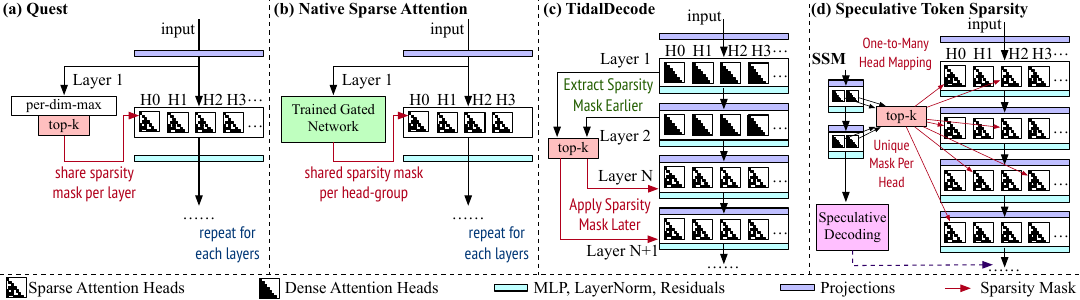}
    \caption{\name vs. Other Token Sparsity Works. The key difference is what ``proxy'' is used for selecting the important tokens. }
    \label{fig:tk_sparse_works}
\end{figure*}

\subsection{Linear Attention and Token Pruning}
\label{sec:lin_attn_tk_prune}



While techniques like speculative decoding can mitigate the latency of auto-regressive generation, the underlying $O(N^2)$ complexity of the attention mechanism remains the primary bottleneck for scaling LLMs to long context windows. 
To circumvent this quadratic complexity, two prominent strategies that perform lossy context state compression have emerged: linear attention and token pruning. 
Linear attention methods, such as Linformer~\cite{wang2020linformer}, Performer~\cite{choromanski2020performer},  Reformer~\cite{kitaev2020reformer}, and Mamba~\cite{gu2024mamba} approximate the attention softmax operation with transformations that have linear time and space complexity. 
Token pruning, employed by methods like H2O~\cite{zhang2023h2o} and \textbf{Scissorhands}~\cite{liu2023scissorhands}, is a more direct approach that compresses or permanently discards tokens from the key-value (KV) cache based on specific heuristics, while keeping the core attention computation unmodified.

Both strategies, however, share a fundamental limitation: they perform irreversible, lossy compression of the context. 
As articulated in multiple related works~\cite{zhang2023h2o,tang2024quest,xiao2023streamingllm,yuan2025nsa}, the importance of a token is dynamic, and it is impossible to know what information will be critical for future queries. 
This limitation is particularly acute in multi-round agentic workloads, as illustrated in Figure~\ref{fig:linear_prune_sparsity}(a) and (b). 
In these scenarios, the model must decide what context to discard or compress in one turn, without knowing what information will be required to respond to environmental feedback in a subsequent turn. 
Prematurely removing a token that later becomes crucial can lead to a catastrophic loss of model capability, especially given the "Lost in the Middle" phenomenon~\cite{liu2024lost} where critical information often resides in the middle of the context.

Nevertheless, the appeal of these methods lies in their potential to optimize not only the attention computation but also the size of the KV cache itself—a key challenge in long-context serving~\cite{kwon2023efficient}.
However, because maintaining a lossless context is essential for robust performance, the risk of severe accuracy degradation makes these approaches ill-suited for complex, unpredictable applications. This motivates the exploration of token sparsity, a lossless alternative.

\subsection{Token Sparsity}
\label{sec:token_sparsity}
Token sparsity presents a more promising alternative that preserves the full, lossless KV cache. 
Instead of compressing the context, these methods exploit the natural sparsity of the attention matrix~\cite{child2019generating,beltagy2020longformer,zaheer2020big} by computing attention on a dynamically selected, small subset of important tokens for each query. 
This approach maintains the full context for future steps while reducing the computational cost of the current step. 
The central challenge, however, lies in efficiently and accurately identifying this critical subset of tokens. 

Current methods for token sparsity mainly differ in their strategies for token selection, often depending on a "proxy" representation to assess the significance of tokens.
We provide a graphical illustration of several approaches in Figure~\ref{fig:tk_sparse_works} and a qualitative comparison of their trade-offs in Table~\ref{table:method_comparison}.

\textbf{Heuristic-Based Selection.}
Some methods identify important tokens using runtime heuristics~\cite{liu2023deja,tang2024quest}. 
Quest, depicted in Figure~\ref{fig:tk_sparse_works}(a), groups tokens into blocks and computes a compact ``summary vector'' for each block through channel-wise Min/Max. 
During generation, these summary vectors act as a proxy to efficiently identify the most relevant blocks of the KV cache. 
Similarly, TidalDecode~\cite{yang2024tidaldecode}, shown in Figure~\ref{fig:tk_sparse_works}(c), posits that attention patterns from early model layers can predict token importance for later layers. 
It computes dense attention for the first few layers and leverages the resulting scores as a heuristic to sparsify computation in subsequent layers.

\textbf{Trained Selection.}
An alternative approach involves training a dedicated gating network to select important tokens, rather than relying on hand-crafted heuristics. 
For instance, DeepSeek's Native Sparse Attention (NSA)~\cite{yuan2025nsa}, shown in Figure~\ref{fig:tk_sparse_works}(b), uses a trained gating network to generate sparsity masks. 
While this can yield highly accurate masks, it requires extensive model retraining. 
This makes the approach impractical for deploying on existing, pre-trained dense models and infeasible for developers without access to large-scale computational resources.

\textbf{Performance Considerations.}
A significant drawback of these approaches is the selection overhead they introduce. 
The heuristic computations or gating network inferences are performed just-in-time, adding latency that can offset the performance gains from sparsity. Furthermore, a critical limitation, summarized in Table~\ref{table:method_comparison}, is that many existing methods like Quest and TidalDecode support sparsity only during the decode phase, requiring a dense prefill. 
This design is fundamentally incompatible with multi-turn agentic workloads for two reasons.

First, agentic workloads are often prefill-bound; the input context from the environment can be long, while the generated output (e.g., a tool call) may be short. 
A decode-only optimization thus fails to accelerate the most expensive phase. 
Second, the multi-turn nature of agentic interactions involves a continuous cycle of prefill (processing new information) and decode (generating a response). 
A decode-only sparse method creates a state mismatch: the sparsely computed KV cache from one turn's decode step cannot be used by the dense prefill required in the next turn. 
This incompatibility effectively breaks inter-turn KV caching, forcing costly recomputation and negating a primary benefit of context management in multi-turn sessions. 
Our work, \name{}, is designed to overcome these limitations.

\section{Motivation}

In recent years, agentic workflows have emerged as a central serving use case, consuming a vast majority of inference tokens due to their iterative, autonomous nature~\cite{ReAct, liu2024agentbench}. 
However, serving these workloads efficiently poses distinct challenges that existing methods fail to address.
While STS introduces unique architectural features that benefit any standard LLM infrastructure, its design is particularly advantageous for serving as a model for \textit{agentic applications}. 

\textbf{The Challenges of Agentic Serving: }
Unlike simple chat applications, agentic workloads typically involve longer sequence lengths and often exhibit more complicated request patterns. 
An agent does not merely generate text; it engages in a continuous loop of reasoning and interacting with the environment. 
This results in a cycle involving long initial prefills, ``prefill-after-decode'' (processing environmental feedback), and multi-turn conversations that require KV caches to persist for extended periods. 

Furthermore, these systems are highly \textit{latency-critical}. 
An agent must wait for the LLM's decoding to finish before it can further interact with the environment. 
Consequently, the serving latency directly dictates the timeliness of the agent's actions and the overall task-level timeliness.


Finally, the need for long sequence lengths in agentic applications requires massive amounts of memory for KV-cache storage. 
In practice, such constraints make \textit{KV-cache offloading} a necessity as agents often need to wait for environment responses (e.g., shell commands)~\cite{FlexGen}, allowing the system to serve other users in the interim. 
However, because it is impossible to keep all active KV caches in GPU memory, the system must frequently swap contexts between CPU and GPU, making the latency caused by swapping a significant concern.

\textbf{The Failure of Existing Methods: }
Current optimization techniques fail to meet these demands in terms of accuracy, capabilities, and performance.

First, methods relying on \textit{lossy context} (e.g., token pruning or compression) are effectively unusable for agents. 
In an autonomous workflow, it is impossible to predict which specific piece of historical information the agent will need in a future step. 
Proactively pruning context is dangerous, as it often leads to irreversible information loss that damages task-level accuracy and reasoning capabilities.

Second, existing lossless sparse attention methods typically support sparsity only during the decoding phase, with their \textit{prefill stages remain 100\% dense}. 
While this may suffice for single-turn chat, it is catastrophic for multi-turn agentic loops. 
In a multi-turn environment, the KV cache generated by a [Dense Prefill] $\rightarrow$ [Sparse Decode] sequence in turn $N$ must be reused in turn $N+1$. However, because these frameworks do not support sparse prefill, the system cannot execute a sequence of [Sparse Decode] $\rightarrow$ [Sparse Prefill] $\rightarrow$ [Sparse Decode]. 
Instead, the context generated sparsely in previous turns must be re-processed densely when new prefill requests arrive (e.g., appending tool outputs). This constant re-computation offsets the gains from sparse decoding, failing to alleviate the raw compute and memory bottlenecks.

\textbf{The STS Solution: }
STS is designed to bridge this gap by decoupling mask generation from the target model's execution. 
By utilizing a draft model to generate sparsity masks asynchronously, STS eliminates latency overhead on the critical path of the target model inference, thereby meeting the strict demands of agentic serving.

Crucially, STS provides the \textit{``known-in-advance''} property. 
Unlike methods that compute the sparsity masks just-in-time, STS determines the sparsity pattern before the target model begins computation. 
This provides a unique opportunity to \textbf{prefetch offloaded KV-caches}. 
Moreover, STS is training-free and targets practical deployment, naturally integrating into existing vLLM infrastructures and speculative decoding frameworks to provide a robust solution for the next generation of AI agents.
Finally, as we will show in the following sections, \name achieves the state-of-the-art sparsity, model-level accuracy, and performance at the same time.

\section{\name Design}\label{sec:sts_algo}

\begin{figure}
  \centering
  \includegraphics[width=0.98\linewidth]{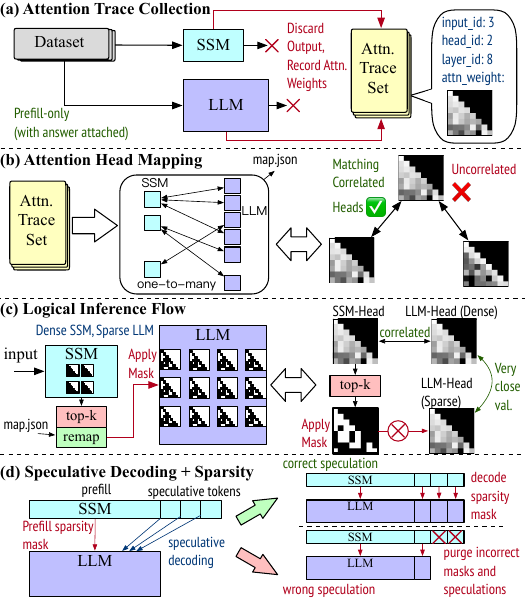}
  \caption{Overview of the \name Workflow. }
  \label{fig:sts_flow}
\end{figure}

This section presents the algorithmic details of \namens. 
The overall workflow, illustrated in Figure~\ref{fig:sts_flow}, consists of a one-time offline head mapping phase followed by an online inference phase. 
We organize our discussion around these stages.

The first stage involves generating a static mapping between the attention heads of the smaller draft model and those of the larger target model. 
This process itself has two steps. 
We first collect a set of attention weight traces from both models on a representative dataset, as shown in Figure~\ref{fig:sts_flow}(a). 
We then use these traces to compute a correlation-based mapping, pairing each target model head with a draft model head as depicted in Figure~\ref{fig:sts_flow}(b). 
The algorithms for trace collection and head mapping are detailed in Section~\ref{sec:head_map}.

Once the mapping is generated, \name performs sparse inference. 
We present the inference flow in two parts for clarity. 
First, in Section~\ref{sec:sts_logical_inference}, we describe the logical flow of computation, explaining how the draft model's attention scores are used to create sparsity masks for the target model (Figure~\ref{fig:sts_flow}(c)). 
Second, in Section~\ref{sec:sts_speculative}, we explain how \name integrates into a speculative decoding system, detailing the necessary mechanisms to handle the asynchronous execution of the draft and target models (Figure~\ref{fig:sts_flow}(d)).

\subsection{Attention Head Mapping}\label{sec:head_map}

A central challenge in using a small model to guide a large one, an approach inspired by speculative decoding, is the mismatch between the two.
Differences in hidden dimensions and the number of layers make it challenging to establish a direct correspondence between their internal representations. 
The \name algorithm overcomes this by leveraging a key similarity of the Transformer architecture: the attention mechanism. 
While models vary in size, the attention weight matrix for any head, regardless ofwhether it comes from the small draft model or the large target model, is consistently shaped $N \times N$, where $N$ is the sequence length. 
A model's size only determines the number of attention heads, with larger models having more than smaller ones, but does not change the $N \times N$ nature of the attention. 
This insight simplifies the problem from one of incompatible shapes to one of mapping a smaller set of draft model heads to a larger set of target model heads.
\name can therefore establish this mapping based on the correlation between the heads' attention patterns.

The first step in creating this mapping is an offline trace generation phase, as illustrated in Figure~\ref{fig:sts_flow}(a). 
The trace generation phase requires a small \textit{trace generation dataset} to be prepared.
During trace generation, we perform a prefill-only forward pass for each sample through both the draft and target models. 
For each forward pass, we record the attention weight matrices from each head in both models and store them in a trace file for the subsequent mapping step.
To demonstrate the generalizability of our approach, we use only the first 100 samples from the WikiText-2 dataset~\cite{merity2016pointer} as our \textit{trace generation dataset} for all evaluations in this paper. 

\begin{algorithm}[t]
\caption{STS Head Mapping Construction}
\label{alg:head_mapping}
\begin{algorithmic}[1]
\Procedure{FindHeadMapping}{$A_D, A_T, k$}
\\ \Comment{$A_D, A_T$: Attention traces from the draft and target.}
\\ \Comment{$k$: The number of top attention scores to select.}
    
    \State $H_D \gets \text{Heads}(A_D)$ \Comment{Set of heads in the draft model}
    \State $H_T \gets \text{Heads}(A_T)$ \Comment{Set of heads in the target model}
    \State $\mathcal{M} \gets \emptyset$ \Comment{Initialize an empty map}

    \For{each target head $h_t \in H_T$}
        \State $max\_score \gets -1$
        \State $best\_match \gets \text{NIL}$
        \State $M_t \gets \text{TopKIndices}(A_T(h_t), k)$

        \For{each draft head $h_d \in H_D$}
            \State $M_d \gets \text{TopKIndices}(A_D(h_d), k)$
            \State $score \gets |M_t \cap M_d|$ \Comment{Compute the total number of matching indices of the two heads}

            \If{$score > max\_score$}
                \State $max\_score \gets score$
                \State $best\_match \gets h_d$
            \EndIf
        \EndFor
        \State $\mathcal{M}[h_t] \gets best\_match$
    \EndFor
    \State \Return $\mathcal{M}$
\EndProcedure
\end{algorithmic}
\end{algorithm}

Once the attention traces are collected, the next step is to construct a mapping from the heads of the target model to those of the draft model. Because the target model has more attention heads than the draft model, this process results in a one-to-many mapping, where a single draft head may be responsible for generating the sparsity mask for multiple target heads. 
The objective is to pair heads that exhibit highly correlated attention patterns, as shown in Figure~\ref{fig:sts_flow}(b). 
While a standard metric like the Pearson correlation coefficient could measure the similarity of the traces, it aligns sub-optimally to our optimization target: 
\name's effectiveness hinges on how well the draft model's `top-k` attention scores predict the target's `top-k` scores. 
Therefore, the ideal correlation metric must directly evaluate the similarity of the resulting token masks, not the raw attention values. 
Based on this reasoning, we developed the procedure detailed in Algorithm~\ref{alg:head_mapping}.

Our algorithm iterates through each attention head in the target model and finds the best-matching head from the draft model. 
The quality of a match is determined by the similarity of the sparsity masks they generate across the collected traces. 
Specifically, for a given target head, the algorithm identifies the draft head whose `top-k` indices produce the largest intersection with the oracle `top-k` indices from the target head. 
It is important to note that this process yields a unique mapping for each value of the sparsity parameter $k$. 
Since the mapping algorithm is run offline on a small trace set, it is computationally cheap and introduces no online overhead.

\subsection{Inference Flow of \namens}\label{sec:sts_logical_inference}

At runtime, \name leverages the draft model to guide the target model's sparse attention computation based on the head mappings already prepared. This process is illustrated in Figure~\ref{fig:sts_flow}(c). 
The inference flow begins with a standard, dense forward pass of the input sequence through the draft model. 
From this pass, the attention weights of each head are extracted. A `top-k` operation is then applied to these weights to identify the indices of the most important tokens, creating a sparsity mask for each draft head. 
Subsequently, these masks are remapped using the offline-generated mapping dictionary, which assigns the correct sparsity indices to each corresponding head in the target model. 
Finally, the target model performs its forward pass, with its attention layers computing sparsely by attending only to the key-value pairs specified by the provided indices.
The sparsity indices are generated at the token level, enabling fine-grained token-wise sparse attention. By applying top-$k$ selection over blocks instead of individual tokens, \textsc{STS} can also support block-wise sparsity, allowing it to adapt to diverse deployment targets.

A key feature of \name is its applicability to both the prefill and the decode stages. 
The core logic remains identical, but the structure of the attention mask differs based on the computation being performed. 
During prefill, multiple query tokens attend to the input sequence, resulting in a two-dimensional attention pattern for each head. 
During the decode stage, a single new token attends to the entire key-value cache. 
Consequently, the draft model generates a one-dimensional sparsity mask for each head, indicating the most relevant tokens in the context for that single new query.




\subsection{\name Integration into Speculative Decoding}\label{sec:sts_speculative}

Speculative decoding executes two models in tandem: a lightweight draft model proposes a bundle of candidate tokens while the heavyweight target model verifies them in one parallel decode step.
\name piggybacks on this infrastructure by treating the draft model as an oracle for guiding the target model's attention sparsity.  
In each speculative round, the draft model runs over the speculated context and emits the attention logits that the STS controller converts into sparsity masks.  
When the target model begins its verification pass, it uses that mask to perform sparse attention.
After the target finishes its verification pass, speculative decoding either accepts a prefix of the candidates or rejects them all.  \name handles the two cases as follows:

\begin{itemize}[leftmargin=*]
    \item \textbf{Successful speculation: }  When the target model accepts a chunk of speculatively decoded tokens, the corresponding sparse metadata has already been consumed during verification.  No further action is required with the KV-cache for the accepted tokens continue to persist in the memory, and the runtime simply advances to the next round. 
    \item \textbf{Speculation failure:}  If the target rejects the proposal (for example, due to a mismatch on the first token), the runtime discards both the speculative tokens and the sparsity mask that was paired with them. The next speculative attempt triggers a fresh draft forward pass, generating a new mask that reflects the updated context.
\end{itemize}



This coupling keeps \name strictly aligned with speculative decoding: masks are generated exactly once per speculative batch, consumed immediately by the target verification pass, and either retained implicitly through accepted KV cache states or thrown away with the rejected bundle.  
The asynchronous execution of the draft and target runners is bridged by the runtime described in Section~\ref{sec:sts_impl}, allowing sparse attention to remain functional even when the speculative verifier backtracks.

\section{\name Implementation}\label{sec:sts_impl}
We implement \name on top of vLLM~\cite{kwon2023vllm} (v0.9.2) and FlashInfer~\cite{ye2024flashinfer} (v0.1.6). Our implementation comprises approximately 7,000 lines of Python and 2,400 lines of CUDA/C++.

\paragraph{Attention Kernels}
We extend FlashInfer's attention kernels to support arbitrary sparsity patterns at block granularity. Each attention head can be assigned with a different set of KV blocks to attend to, and is allocated to a thread block for computation. Split-K trick is applied for longer sequences to improve SM utilization~\cite{ye2024flashinfer}. The KV blocks scattered across the GPU global memory are first read into contiguous shared memory tiles. Attention computation is performed over the shared memory and is pipelined with shared memory loading. As STS computes the LLM's attention masks using the attention scores of the LLM, we also modify the full attention kernel to output the product between queries and keys into a pre-allocated buffer in global memory.

\paragraph{Drafting-Speculation Pipelining}
To minimize the overhead of generating the sparsity mask, we implement a pipeline execution mechanism where the forward of layer $i$ of the SSM is overlapped with the computation of the sparsity masks of the LLM's attention heads that are mapped to that from layer $i-1$ from the SSM. The forward computation and mask generation are placed on separate CUDA streams, while their data dependency is resolved with CUDA event synchronization. The mask computation fully resides on GPU and is compatible with CUDA graphs for efficient execution.


\section{Evaluation}\label{sec:eval}
In this section, we evaluate \name on its performance and accuracy.
Our evaluation addresses three critical questions:
\begin{itemize}[leftmargin=*]
    \item \textbf{Performance (Section~\ref{sec:system_perf}):} How does the theoretical sparsity degree of \name effectively translate into wall-clock speedups. How does the ``know-in-advance'' feature of \name help hide the latency during KV-cache offloading?
    \item \textbf{Fidelity \& Robustness (Section~\ref{sec:fidelity}):} Does \name{} preserve generation quality across diverse workloads, ranging from "needle-in-a-haystack" retrieval to complex reasoning?
    \item \textbf{Insights from Micro-Analysis (Section~\ref{sec:micro}):} How does the draft-guided sparsity pattern align with the actual attention mechanisms of Large Language Models?
\end{itemize}

\subsection{Experimental Setup}
\label{sec:setup}

\noindent\textbf{Hardware Testbed.}
All experiments are conducted on a high-performance server equipped with NVIDIA H20 GPUs~(96GB HBM3 memory). 
To rigorously evaluate our offloading mechanism, the GPU is connected to the host CPU via a PCIe Gen5x16 interface, providing a theoretical bi-directional bandwidth of 128 GB/s. 
This setup represents a typical production environment for serving large-scale models.

\vspace{0.5em} 
\noindent
\textbf{Software \& Baselines.} 
We compare against a comprehensive suite of baselines:
\begin{itemize}[leftmargin=*]
    \item \textbf{Dense (Full Attention):} The gold standard for accuracy, but computationally expensive ($O(N^2)$).
    \item \textbf{Quest~\cite{tang2024quest}:} Leverages query-aware sparsity for efficient long-context inference. It identifies critical tokens by estimating the importance of KV cache pages relative to the current query.
    \item \textbf{TidalDecode~\cite{yang2024tidaldecode}:} A method that selects tokens based on spatial locality and sparse metrics.
    \item \textbf{StreamingLLM~\cite{xiao2023streamingllm}:} A static sparsity baseline that keeps only the initial (sink) and recent tokens.
\end{itemize}
We use \textbf{Llama-3.1-8B-Instruct} as the target model and \textbf{Llama-3.2-1B-Instruct} as the draft model. 
Unless stated otherwise, experiments use greedy decoding (temperature=0) to ensure deterministic reproducibility.

\subsection{End-to-End System Performance}
\label{sec:system_perf}
\begin{figure}[t]
  \centering
  
  \begin{subfigure}[b]{\linewidth}
    \centering
    \includegraphics[width=0.95\linewidth]{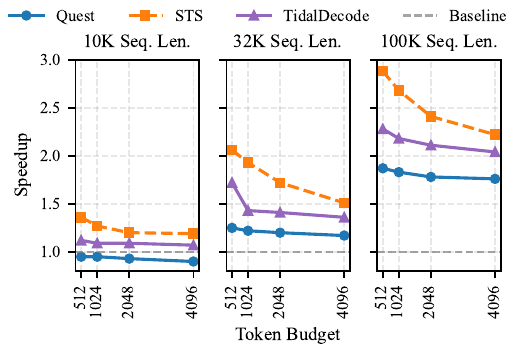}
    \caption{Speedup across varying sequence lengths}
    \label{fig:speedup_over_len} 
  \end{subfigure}
  
  \vspace{0.5cm} 
  
  \begin{subfigure}[b]{\linewidth}
    \centering
    \includegraphics[width=0.95\linewidth]{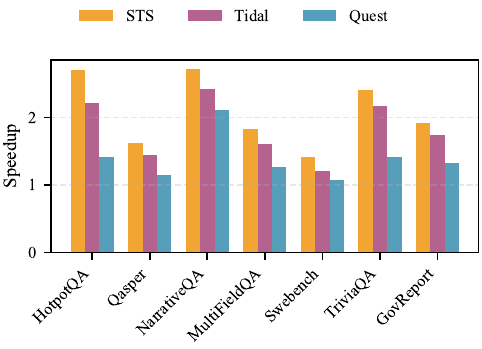}
    \caption{Speedup on real-world workloads}
    \label{fig:speedup_over_ds} 
  \end{subfigure}
  
  \caption{\textbf{[System Performance]: End-to-end speedup analysis.} 
  (a) \name{} scales efficiently up to 100K context length. 
  (b) \name{} maintains performance advantages on realistic benchmarks like LongBench and SwBench.}
  \label{fig:system_perf_combined} %
  
\end{figure}

\subsubsection{End-to-End System Speedup Analysis}
Figure~\ref{fig:speedup_over_len} and Figure~\ref{fig:speedup_over_ds} present the speedup of \name{} relative to the Dense baseline.

\noindent\textbf{Scaling with Context Length.}
As illustrated in Figure~\ref{fig:speedup_over_len}, \name{} demonstrates superior scalability as the sequence length grows from 10K to 100K.
Even at a relatively short context of 10K, \name{} already yields a 1.36$\times$ speedup over the dense baseline, outperforming other sparse attention baselines.
This advantage becomes more pronounced at 32K context, where the speedup rises to 2.06$\times$.
The gap widens significantly in the extreme setting of 100K context length; with a token budget of 512, \name{} achieves a remarkable \textbf{2.88$\times$ speedup}.
In contrast, Quest and TidalDecode plateau at lower speedups as the sequence length increases. This widening gap stems from the overhead of "Just-in-Time" selection.
Methods like Quest require loading coarse-grained Q and K metadata from HBM to compute importance scores \emph{during} the target model's execution. As context length grows (e.g., from 10K to 100K), this memory bandwidth consumption becomes a bottleneck.
\name{}, however, decouples mask generation: the sparsity mask is pre-computed by the draft model. This allows the target model to execute sparse attention kernels immediately without stalling for metadata analysis, ensuring consistent speedup gains across all context lengths.

\noindent\textbf{Real-world Workloads.}
We further evaluated \name{} on realistic datasets, including LongBench and SweBench~\cite{jimenez2024swebench}, as shown in Figure~\ref{fig:speedup_over_ds}.
For these experiments, we standardized the active token budget to \textbf{4096}.
This choice is empirically grounded; our subsequent accuracy benchmarks (see Section~\ref{sec:fidelity}) confirm that a 4096 budget is sufficient to preserve generation quality for these workloads.

Under this setting, \name{} consistently delivers the highest speedup.
We highlight our evaluation methodology on \textbf{SweBench}, which involves complex code generation tasks.
Instead of using synthetic inputs, we adopted a trace-driven approach: we sampled 100 instances from the dataset, collected their execution traces to determine the average sequence length and output token counts, and measured the speedup based on these realistic profiles.
Results indicate that \name{} outperforms TidalDecode by a significant margin on SweBench.
This suggests that \name{}'s draft-based selection is far more robust to the irregular and long-range attention patterns typical of code repositories compared to locality-biased heuristics.
\begin{figure}[p] 
  \centering
  
  \includegraphics[width=0.95\linewidth]{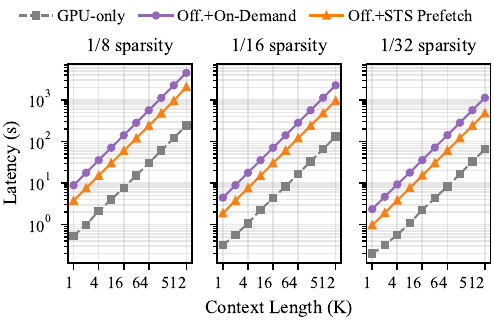}
  \caption{\textbf{[System Optimization]}: Latency of KV-cache offloading strategies. }
  \label{fig:offload}
  
  \vspace{0.5cm} 
  
  \includegraphics[width=\linewidth]{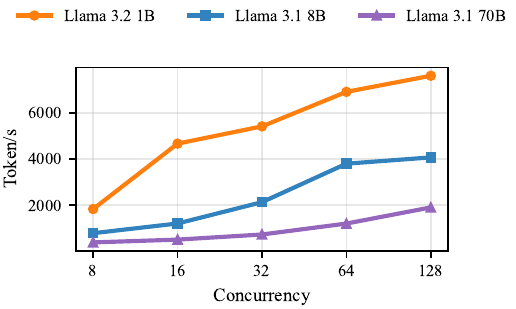}
  \caption{\textbf{[Draft Overhead Analysis]}: Throughput scalability under concurrency.}
  \label{fig:concurrency_vs_tokens}
  
  \vspace{0.5cm} 
  
  \includegraphics[width=0.99\linewidth]{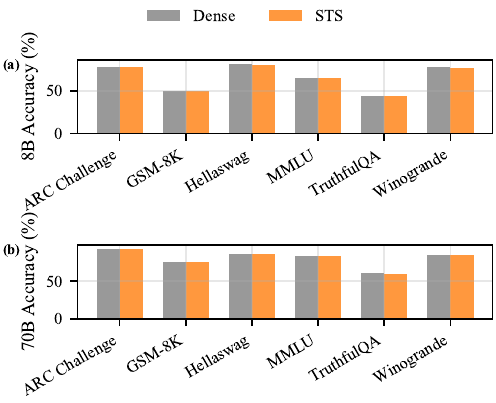}
  \caption{\textbf{[General Benchmark Fidelity]: Zero-shot accuracy on standard reasoning benchmarks.} \name{} incurs negligible accuracy loss across 8B and 70B models, demonstrating that sparsity patterns generalize effectively across model scales.}
  \label{fig:dense_vs_sts_8b70b}
  
\end{figure}
\subsubsection{Hiding Latency via Prefetching}
\label{sec:prefetch}
A critical bottleneck in long-context serving is GPU memory capacity. When the KV cache exceeds HBM limits, systems must offload data to CPU RAM. Standard offloading incurs severe latency penalties due to the PCIe bandwidth bottleneck. 
With the ``know in advance'' feature of \name{}, part of the latency from offloading can be hidden through prefetching. 
Since the draft model runs ahead of the target model, we know exactly which KV blocks will be needed \emph{before} the target model begins its layer computations. 
We leverage this lookahead to issue asynchronous prefetch requests over the PCIe Gen5 bus, transferring blocks from CPU memory to GPU memory before they are strictly needed. 
In memory-constrained scenarios where the full KV cache cannot fit in GPU memory, this prefetching capability significantly reduces the performance overhead of CPU offloading.

We evaluate the effectiveness of prefetch-enabled KV-cache offloading by comparing three strategies:
\begin{itemize}[leftmargin=*]
    \item \textbf{Full GDDR:} The entire KV-cache resides in GPU memory, serving as the ideal performance baseline.
    \item \textbf{On-Demand:} KV-cache blocks are stored in CPU memory and transferred to the GPU synchronously only when needed, blocking computation.
    \item \textbf{Prefetch (\namens):} \name{} predicts the required KV-cache blocks several tokens ahead and initiates asynchronous transfers that overlap with computation.
\end{itemize}

Figure~\ref{fig:offload} presents the latency comparison across varying context lengths (1K--512K tokens) and sparsity levels (1/8, 1/16, and 1/32). The \textbf{On-Demand} approach incurs substantial overhead, with latency increasing by 14--19$\times$ compared to the GPU-only baseline due to synchronous memory transfers. In contrast, the \textbf{Prefetch} approach reduces this overhead to 6--8$\times$ by effectively overlapping data transfers with kernel execution. This represents a roughly \textbf{3$\times$ reduction} in effective latency. These results demonstrate that \name{}'s ability to predict future sparse attention patterns enables effective hiding of memory transfer latency, making CPU offloading a viable option for serving long-context workloads when GPU memory is limited. From a resource utilization perspective, this strategy effectively converts the idle PCIe bandwidth during the computation phase into useful data throughput. By saturating the bus asynchronously, we prevent the I/O subsystem from becoming the critical path, achieving a "compute-bound" rather than "memory-bound" execution profile.

\subsubsection{Scalability and Overhead Analysis}
To assess the system's robustness in real-world serving scenarios, we evaluate the throughput scalability of \name{} under varying concurrency levels.

\noindent\textbf{Draft Model Efficiency.}
A potential concern with draft-based speculation is the computational overhead of the draft model itself. 
Figure~\ref{fig:concurrency_vs_tokens} dispels this concern by comparing the throughput of the draft model (Llama-3.2-1B) against the target models (Llama-3.1-8B/70B).
The draft model achieves orders of magnitude higher throughput (e.g., $>6000$ tokens/s at concurrency 128). 
This extreme speed ensures that the mask generation step accounts for a trivial fraction of the total end-to-end latency.

\noindent\textbf{Scalability under Load.}
As concurrency increases from 8 to 128, the draft model exhibits linear scalability without saturation.
This indicates that \name{} is well-suited for high-throughput serving environments, where the draft model can efficiently service multiple target model instances without becoming a bottleneck.
\begin{figure*}[t]
  \centering
  \includegraphics[width=0.99\textwidth]{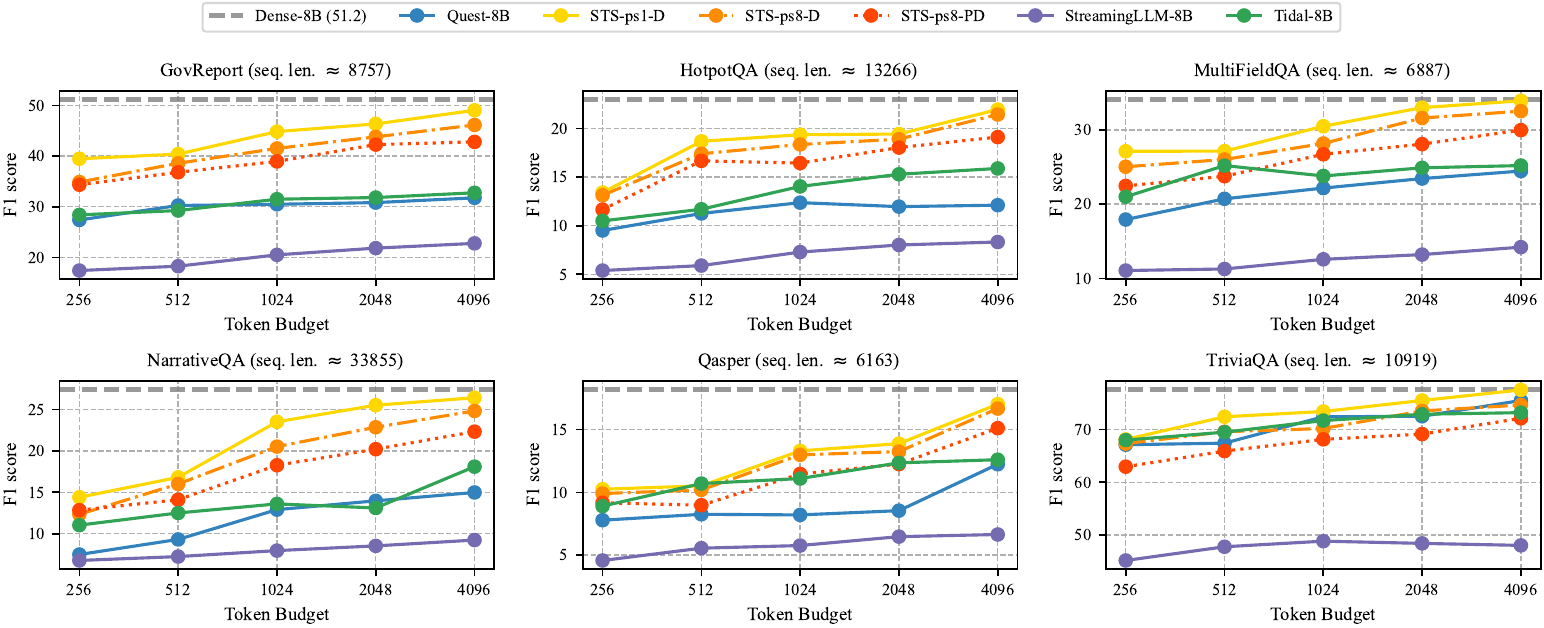}
  \caption{\textbf{[LongBench Fidelity]: LongBench results on Llama-3.1-8B.} \name{} (red) consistently outperforms sparse baselines (Quest, Tidal) and matches Dense performance across diverse tasks.}
  \label{fig:longbench_8b}
\end{figure*}

\begin{figure*}[t]
  \centering
  \includegraphics[width=0.99\textwidth]{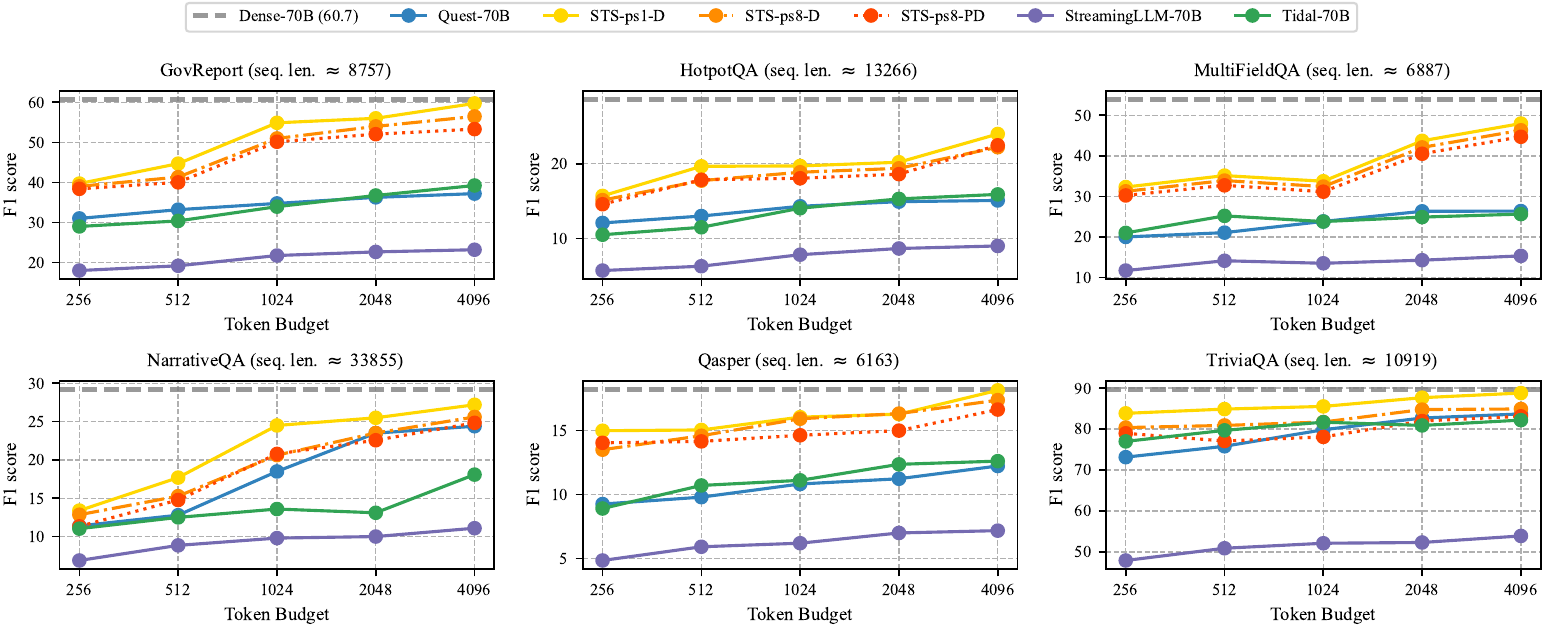}
  \caption{\textbf{[LongBench Fidelity]: LongBench results on Llama-3.1-70B.} \name{} scales effectively to larger models, preserving long-range dependencies even in multi-document QA tasks.}
  \label{fig:longbench_70b}
\end{figure*}
\subsection{Fidelity and Robustness}
\label{sec:fidelity}
Speed is irrelevant if it is at the cost of destroying the model's accuracy. We extensively verify that \name{} maintains the comparable quality of the dense baseline. Before analyzing the results, we define the sparsity scopes and granularity configurations used in our evaluation:
\begin{itemize}[leftmargin=*]
    \item \textbf{Sparse Decode (STS-D) vs. Sparse Prefill (STS-P).} 
    We evaluate two configurations: (1) \textbf{STS-D}, where the full KV cache is preserved during the prefill phase, and sparsity is applied only during token generation; and (2) \textbf{STS-P}, a stricter setting where the prompt itself is compressed on-the-fly, retaining only critical blocks in memory. STS-P serves as a stress test for the draft model's ability to identify semantic importance without hindsight.
    
    \item \textbf{Page Size Granularity ($p_s$).} 
    Since vLLM and FlashInfer manage memory in paged blocks, Page Size ($p_s$) is a critical factor. A smaller $p_s$ allows for finer-grained selection but incurs higher indexing overhead. 
\end{itemize}
\subsubsection{Standard \& Long-Context Benchmarks}

\noindent\textbf{General Reasoning (8B \& 70B).} 
Figure~\ref{fig:dense_vs_sts_8b70b} shows zero-shot accuracy on benchmarks like MMLU~\cite{hendrycks2020measuring} (knowledge), GSM8K~\cite{cobbe2021training} (math), and ARC~\cite{clark2018think} (reasoning).
\name{} tracks the Dense baseline within 0.5\% variance across all tasks for both 8B and 70B models. 
\emph{Implication:} This result is significant because it proves that the sparsity patterns learned by a small 1B model are semantically aligned with the reasoning pathways of a 70B model. The "critical thinking" tokens are shared across model scales.

\noindent\textbf{Long-Context Robustness \& Efficiency.} 
We evaluate performance on LongBench~\cite{bai2023longbench} across Llama-3-8B (Figure~\ref{fig:longbench_8b}) and Llama-3-70B (Figure~\ref{fig:longbench_70b}). 
\name{} demonstrates superior robustness compared to baselines, maintaining high fidelity even under constrained conditions.

\begin{itemize}[leftmargin=*]
    \item \textbf{Resilience at Low Budgets:} 
    As shown in the figures, heuristic baselines like Quest and Tidal suffer sharp performance drops when the token budget is restricted (e.g., 256 or 512 tokens in HotpotQA). In contrast, \name{} maintains a much flatter degradation curve. This indicates that our draft-based selection effectively identifies the "Pareto set" of critical tokens, preserving essential information where other methods fail.
    
    \item \textbf{Granularity \& Prefill Viability:} 
    Crucially, our performance advantage does not rely on expensive fine-grained access. 
    The results show that the coarse-grained configuration ($P_s=8$, blue line) retains nearly 98\% of the performance of the fine-grained setting ($P_s=1$, red line). This confirms that semantic information is locally clustered, allowing us to exploit hardware-friendly block-sparse kernels.
    Furthermore, the method remains effective even in the stricter \textbf{STS-PD} setting (sparse prefill, green line), where non-essential tokens are pruned during prompt processing. The fact that STS-PD performs comparably to the standard decode-only sparsity suggests that \name{} can safely compress prompts on-the-fly without losing context.
\end{itemize}

\begin{figure}[t]
  \centering
  \includegraphics[width=0.99\linewidth]{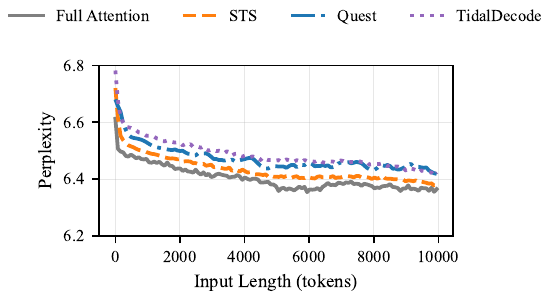}
  \caption{\textbf{[Fidelity]: Language modeling perplexity on Wiki-103.} \name{} maintains perplexity within 0.01 of the dense baseline, whereas heuristic methods degrade at long contexts.}
  \label{fig:exp3_ppl_length}
\end{figure}

\begin{figure}[t]
  \centering
  \includegraphics[width=0.99\linewidth]{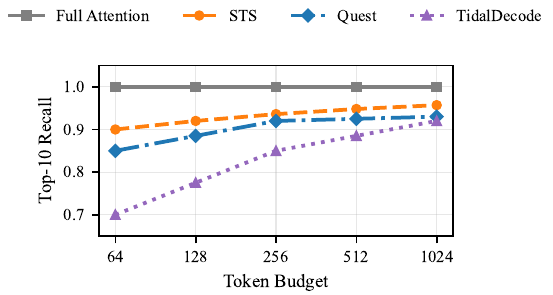}
  \caption{\textbf{[Fidelity]: Passkey retrieval recall.} \name{} maintains >98\% recall even at high sparsity levels, proving effective preservation of critical evidence tokens.}
  \label{fig:exp2_passkey}
\end{figure}

\begin{figure}[t]
  \centering
  \includegraphics[width=0.99\linewidth]{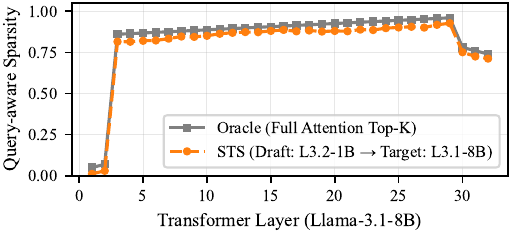}
  \caption{\textbf{[Micro-analysis]: Layer-wise sparsity patterns.} The draft-guided mask (STS) closely aligns with Oracle's distribution.}
  \label{fig:exp1_sparsity}
\end{figure}
\subsubsection{Synthetic Probes: PPL and Passkey}
\noindent\textbf{Perplexity Stability.} 
Figure~\ref{fig:exp3_ppl_length} confirms that \name{} does not drift as context grows. 
While TidalDecode shows increasing perplexity degradation ($\Delta > 0.5$) at lengths beyond 8K, \name{} stays within $\Delta < 0.01$ of the Dense baseline. 
This stability indicates that \name{} correctly attends to long-range linguistic dependencies that are crucial for next-token prediction.
\vspace{0.5em} 
\noindent\textbf{Needle-in-a-Haystack Retrieval.} 
The Passkey task (Figure~\ref{fig:exp2_passkey}) is a stress test for attention mechanisms. 
\name{} achieves \textbf{>98\% recall} with a budget of just 256 tokens out of 10K. 
In comparison, heuristic methods often discard the passkey because it appears in a "low-entropy" position. 
The draft model, however, recognizes the semantic anomaly of the passkey and flags it for the target model to attend to. This phenomenon aligns with the information-theoretic concept of "surprisal." A random passkey inserted into coherent text disrupts the statistical flow, creating a high-entropy signal. Even a small 1B model is sensitive to such disruptions and assigns high attention scores to these anomalies, effectively acting as a "sieve" that catches irregularities for the larger model to process.

\subsection{Micro-Analysis: Layer-wise Sparsity}
\label{sec:micro}

To understand the source of \name{}'s efficiency and efficacy, we perform a detailed analysis of the sparsity patterns it generates. 
Figure~\ref{fig:exp1_sparsity} plots the \textbf{Oracle Prunable Ratio} alongside the mask generated by our method.  We define this Oracle metric as the theoretical upper bound of sparsity: for each layer, it represents the maximum fraction of attention entries that can be discarded without exceeding a strict perplexity degradation threshold ($\Delta\text{PPL}<0.01$). This sensitivity analysis reveals the inherent \textit{heterogeneity} of the model: some layers are "information-dense" and require full attention, while others are highly redundant.

We observe a distinct \textbf{"inverted-U"} sparsity pattern across layers:

\begin{itemize}[leftmargin=*]
    \item \textbf{Initial Layers (Dense):} The first 2 layers require high density (only 5--10\% sparsity). This aligns with the \emph{Attention Sink} hypothesis~\cite{xiao2023streamingllm}, suggesting that initial tokens serve as anchors for global context aggregation. Furthermore, these layers are responsible for low-level feature extraction, requiring broad access to local neighbors to form basic semantic representations.
    
    \item \textbf{Middle Layers (Highly Sparse):} Layers 3--28 tolerate extreme sparsity (85--95\%). These layers typically process local syntax or perform "associative memory" lookups. As noted in induction head theory~\cite{olsson2022context}, such operations are highly selective, focusing on only a few specific tokens (e.g., copying from previous occurrences) while ignoring the vast majority of the context.
    
    \item \textbf{Final Layers (Moderate):} The last few layers exhibit a drop in sparsity (retaining $\sim$30\% of tokens). This indicates a need to synthesize broader context features and calibrate probability distributions for the final next-token prediction, requiring a more holistic view of the sequence.
\end{itemize}

Crucially, the sparsity mask generated by the draft model (STS) closely tracks this Oracle curve. 
This confirms our core hypothesis: \emph{small language models share the same fundamental attention topology as large models}, making them excellent predictors for sparse attention without the need for expensive gradients or full-model computations.

\section{Related Work}\label{sec:rela}
We classify efficient long-context inference into three paradigms: state eviction (KV compression), heuristic approximation (sparse attention), and speculation-assisted systems.

\subsection{KV Cache Compression \& Eviction}
Systems like H2O~\cite{zhang2023h2o}, StreamingLLM~\cite{xiao2023streamingllm}, SnapKV~\cite{li2024snapkv}, and PyramidInfer~\cite{yang-etal-2024-pyramidinfer} alleviate memory pressure by permanently evicting tokens based on heuristics such as attention scores or positional recency. While these strategies effectively exploit the sparsity of attention matrices, they rely on the assumption that past token importance predicts future relevance. Consequently, they enforce permanent state eviction, causing irreversible information loss. This is catastrophic for agentic workflows where historical context—often located in the "long tail" of the attention distribution—becomes relevant unexpectedly during multi-step reasoning.

Alternatively, memory-augmented approaches like InfLLM~\cite{xiao2024infllm} offload KV blocks to CPU memory or disk to avoid data loss. Although this preserves the full context, the limited bandwidth of PCIe interconnects introduces significant data movement overhead. The high latency of retrieving these blocks during the decoding phase creates a bottleneck, preventing the system from meeting the strict latency requirements of real-time interaction.

Unlike these approaches, \name{} offers the capability to maintain a lossless, GPU-resident context by decoupling storage from computation, ensuring both data integrity and high-throughput access.

\subsection{Sparse Attention}
To reduce $O(N^2)$ complexity, existing approaches generally fall into static, dynamic, or learned categories.
Static methods (e.g., MInference~\cite{jiang2024minference}, DuoAttention~\cite{xiao2025duoattention}) rely on pre-defined patterns, while dynamic systems (e.g., Quest~\cite{tang2024quest}, TidalDecode~\cite{yang2024tidaldecode}) estimate token importance on-the-fly. 
However, dynamic heuristics often incur synchronization overheads on the critical path, and simple metrics may miss semantically subtle "needles."

More recently, SeerAttention~\cite{seerattention} introduces a learnable gating mechanism, distilled from dense attention maps to predict block sparsity. 
While effective, it imposes a significant deployment barrier: it necessitates an additional training phase (e.g., on 0.5B tokens) and structural modifications to the model architecture. Furthermore, SeerAttention primarily targets prefill acceleration, leaving the decoding phase dense.

In contrast, \name{} is a \textbf{training-free} solution that requires no architectural changes or distillation. 
It functions as a plug-and-play overlay compatible with existing inference engines (e.g., vLLM), leveraging the draft model to achieve zero-overhead sparsification for both prefill and decoding phases.

\subsection{Speculative Execution and Pipelining}
Speculative decoding~\cite{leviathan2023fast,chen2023accelerating} originally utilized small draft models solely to accelerate token generation. 
Recent works have extended this paradigm to context pruning. For instance, LazyLLM~\cite{fu2024lazyllm} progressively selects tokens required for the next step. 
However, such methods operate on a Just-In-Time (JIT) basis: the selection logic is interleaved with the target model's execution, forcing synchronous waits and preventing kernel overlapping.
\name{} re-architects this into an asynchronous lookahead pipeline. 
By shifting the burden of sparsity planning entirely to the draft phase, we decouple mask generation from target execution. 
This achieves the "known-in-advance" property, enabling system-level optimizations—such as latency hiding via prefetching and pipelined memory accesses—that are unattainable in synchronous JIT designs.

\section{Conclusion}
The quadratic complexity of attention is a critical bottleneck for long-context LLM inference. We introduced STS, a training-free sparse attention mechanism that repurposes the draft model in speculative decoding to generate predictive sparsity masks for the target model. This approach preserves the full KV cache, which is crucial for model fidelity. Our evaluation shows that STS achieves a superior accuracy-sparsity trade-off, maintaining near-lossless performance on language modeling and long-context reasoning benchmarks. By supporting both prefill and decode sparsity, STS is uniquely suited for efficient, multi-turn agentic workloads.

A key contribution of STS is that its sparsity masks are "known-in-advance," enabling novel system-level optimizations like proactive memory prefetching that are infeasible with just-in-time methods. STS thus represents a significant step towards practical and efficient long-context inference, establishing a new state-of-the-art for training-free sparse attention.

\bibliographystyle{plain}
\bibliography{references}

\end{document}